%% file: main.tex
\documentclass[letterpaper, 10 pt,conference]{ieeeconf}  % Comment this line out if you need a4paper

\IEEEoverridecommandlockouts                              % This command is only needed if 
\usepackage[maxnames=99,maxbibnames=99]{biblatex}
\usepackage{hyperref}
\usepackage{subcaption}
\usepackage{svg}
\usepackage{adjustbox}
\usepackage{float}
\usepackage{booktabs}
\usepackage{graphicx}  % for pdf, bitmapped graphics files
\usepackage{epsfig} % for postscript graphics files
\usepackage{times} % assumes new font selection scheme installed
\usepackage{amsmath} % assumes amsmath package installed
\usepackage{amssymb}  % assumes amsmath package installed
\usepackage{mathtools}
\usepackage[T1]{fontenc}
\usepackage{lmodern}   
\renewcommand{\vec}[1]{\boldsymbol{#1}}
\newcommand{\floor}[1]{\left\lfloor #1 \right\rfloor}

\title{\LARGE \bf
Acoustic Drone Package Delivery Detection }

\author{François Marcoux and François Grondin% <-this % stops a space
\thanks{This work is financed through the Fonds de recherche du Qu\'ebec - Nature et Technologies with the Research Support for New Academics grant. F. Marcoux and F. Grondin are with the Department of Electrical Engineering and Computer Engineering, Université de Sherbrooke, Qu\'ebec, Canada}%
}

\begin{document}

\maketitle
\thispagestyle{empty}
\pagestyle{empty}

%%%%%%%%%%%%%%%%%%%%%%%%%%%%%%%%%%%%%%%%%%%%%%%%%%%%%%%%%%%%%%%%%%%%%%%%%%%%%%%%
\begin{abstract}
In recent years, the illicit use of unmanned aerial vehicles (UAVs) for deliveries in restricted area such as prisons became a significant security challenge. 
While numerous studies have focused on UAV detection or localization, little attention has been given to delivery events identification. 
%It has become more important for those vulnerable to illegal UAV delivery to add specialized equipment to detect UAV intrusion and UAV related event. 
This study presents the first acoustic package delivery detection algorithm using a ground-based microphone array. 
The proposed method estimates both the drone's propeller speed and the delivery event using solely acoustic features. 
%is h moment a quadcopter, is more likely to have dropped a package. Paired with sound source localization, it is also possible to mark a zone on the ground were the package was dropped. 
%The quadcopter delivery detection algorithm is split into two steps : acoustic blade passing frequency(BPF) estimation and temporal BPF distribution analysis.
%The algorithms relies on the fact that heavier payload requires higher propeller speed. 
%Since the thrust needed to lift a given weight is directly linked to propeller rotation speed and the acoustic wave generated by those propellers, it is possible to observe weight differences by analyzing sudden change in propellers fundamental frequencies . 
A deep neural network detects the presence of a drone and estimates the propeller's rotation speed or blade passing frequency (BPF) from a mel spectrogram. 
The algorithm analyzes the BPFs to identify probable delivery moments based on sudden changes before and after a specific time. 
%The second step is to compare two  
Results demonstrate a mean absolute error of the blade passing frequency estimator of 16 Hz when the drone is less than 150 meters away from the microphone array. 
The drone presence detection estimator has a accuracy of 97\%. 
The delivery detection algorithm correctly identifies 96\% of events with a false positive rate of 8 \%. 
This study shows that deliveries can be identified using acoustic signals up to a range of 100 meters.

\end{abstract}

%%%%%%%%%%%%%%%%%%%%%%%%%%%%%%%%%%%%%%%%%%%%%%%%%%%%%%%%%%%%%%%%%%%%%%%%%%%%%%%%
\section{INTRODUCTION}
\input{Sections/intro}

\section{Blade passing frequency estimation and delivery detection}
\label{sec:bpf estimation and drop detection}
\input{Sections/drop_detection}

\section{Experimental setup}
\label{sec:Experimental setup}
\input{Sections/experimental_setup}

\section{Results and discussion}
\label{sec:Results and discussions}
\input{Sections/results_and_discussion}

\section{Conclusion}
\label{sec:conclusion}
\input{Sections/conclusion}


%%%%%%%%%%%%%%%%%%%%%%%%%%%%%%%%%%%%%%%%%%%%%%%%%%%%%%%%%%%%%%%%%%%%%%%%%%%%%%%%

%%%%%%%%%%%%%%%%%%%%%%%%%%%%%%%%%%%%%%%%%%%%%%%%%%%%%%%%%%%%%%%%%%%%%%%%%%%%%%%%

%%%%%%%%%%%%%%%%%%%%%%%%%%%%%%%%%%%%%%%%%%%%%%%%%%%%%%%%%%%%%%%%%%%%%%%%%%%%%%%%

%\section*{ACKNOWLEDGMENT}

%%%%%%%%%%%%%%%%%%%%%%%%%%%%%%%%%%%%%%%%%%%%%%%%%%%%%%%%%%%%%%%%%%%%%%%%%%%%%%%%
\renewcommand*{\bibfont}{\normalfont\small}
\printbibliography

\end{document}

%% file: Sections/intro.tex
%Plan de rédaction pour l'intro
% Présentation du contexte : Drone -> utilisé énormément -> problématique dans les no fly zone -> encore plus dans les prisons où ils livrent des paquet -> parler de la tendance de développement de sytème de détection et de localisation. 
% Aucun systeme fait de la détection de moment de largage et ça serait interessant pour facilité l'intervention des gardes

% Présenter les travaus de détection de payload sur un drone et comment il diffère de mon cas d'utilisation

Unmanned aerial vehicles (UAVs) have grown rapidly due to their affordability, maneuverability and long range capabilities.
These features make them well-suited for a variety of applications, including videography, search and rescue missions and item delivery in remote areas. 
However, UAVs pose significant risks in no-fly zones such as airports and prisons.
Authorities report a rise in hostile UAV intrusions over the past few years. %\footnote{\url{https://www.faa.gov/uas/resources/public_records/uas_sightings_report}}
In prisons, UAVs are frequently used to smuggle contraband such as cellphones, weapons, and drugs \cite{ohagan2017behind,norman2023global}. Although active mitigation systems exist and can neutralize drones \cite{dix2021contraband,wang2021counter}, local laws often prohibit radio frequency jamming or drone interception \cite{russo2019countering}. 
Authorities therefore rely on multimodal detection systems to identify and track intrusions in real-time, and eventually recover packages dropped within the perimeter.

Surveys on drone detection systems \cite{park2021survey,besada2022review,sadovskis2022modern} identify four sensor types used for drone surveillance: cameras, radio-frequency antennas, radar, and microphones.
Cameras are relatively inexpensive but their effectiveness is highly dependent on the environment, lighting, and field of view.
Radars provide long-range object detection but are more expensive and susceptible to false positives, as small, slow-moving drones can be mistaken for birds \cite{besada2022review}.
Radio-frequency antenna eavesdrops on data sent between drones and their ground stations. 
Valuable data like GPS and possibly drone delivery state can be extracted.
However, encrypted communication or an autonomous drone with preprogrammed flight missions make this strategy inefficient \cite{sadovskis2022modern}.
Microphone arrays are cost-effective and capable of localizing sound sources, but they have limited range and are sensitive to noise. 
In most scenarios, multimodal systems offer the best overall performance \cite{park2021survey}.

To the best of our knowledge, no prior research addresses package delivery detection.
However, two related tasks have been explored: 1) classification of payload weight \cite{ibrahim2022noise2weight,doster2023uav,ku2022uav,traboulsi2021identification} and 2) detection of payload presence \cite{li2018convolutional,lim2018practically,utebayeva2023practical,wang2022feature}. 
The first studies on the classification of loaded versus unloaded UAVs are presented in \cite{li2018convolutional,lim2018practically}. 
Both use recordings from a DJI Phantom 2 and perform classification with a convolution neural network (CNN) on long audio sequences (5 to 16 seconds).
Mel Frequency Cepstrum Coefficients (MFCC) \cite{li2018convolutional} or Short-Time Fourier Transform (STFT) \cite{lim2018practically} can be used as input features and both achieve more than 99\% accuracy.
Despite their high accuracy, these methods are restricted to a specific drone model and overlook the impact of the distance between the microphone and the drone. 
Other studies aimed to evaluate different machine learning models and feature extraction techniques.
Deep neural networks \cite{ku2022uav} and classical machine learning models \cite{wang2022feature,ibrahim2022noise2weight} have been tested for both payload classification and detection tasks, consistently achieving around 99\% accuracy.
However, data collection is limited to one or two drone models and the audio is recorded at short range (up to 15 meters).
Ibrahim et al. \cite{ibrahim2022noise2weight} and Doster et al. \cite{doster2023uav} further demonstrate that low-cost microphones can capture drone sound up to a range of 100 m, as they achieve 99\% accuracy using MFCC and a Support Vector Machine (SVM) model. 
They also investigate the impact of distance and wind noise level on accuracy. 
Accuracy decreases with distance reaching 70\% at 100 m and wind noise levels above 54 dB significantly impact the accuracy of the predictions. 
Utebayeva et al. \cite{utebayeva2023practical} collect data from 11 different drone models at different locations with multiple payload weights and at distances up to 100 m. 
Using  Mel spectrograms of one second and a Recurrent Neural Network (RNN), they achieved 98\% accuracy. 

Previous work demonstrates that UAV payload detection and classification using acoustic signals is feasible.
However, no studies have focused on identifying the precise moment of a delivery event. 
This paper introduces an acoustic drone delivery detection system. 
We conducted numerous test flights to collect audio data both onboard the drone and from a ground-based station.
We also collect telemetry data, including GPS coordinates and motor pulse width modulated (PWM) control signals.
We use pairs of audio segments and telemetry data to train a supervised convolutional recurrent neural network (CRNN) that estimates the drone activity and the blade passing frequencies (BPF) of the motors. 
Based on the estimated BPF, we design a detection algorithm that predicts delivery events. 

%The contribution of this paper can be resumed with the following: \begin{enumerate}\item We present the first acoustic payload delivery moment identification algorithm;\item We propose a blade passing frequency estimation system that can be used for other UAV acoustic related tasks;\item We present an evaluation of the SNR and drone-to-array distances effect on the accuracy of our approach;\end{enumerate}

The rest of this paper is organized as follow. Section \ref{sec:bpf estimation and drop detection} presents the drone activity and BPF estimation algorithm and the delivery detection algorithm. Section \ref{sec:Experimental setup} describes the experimental setup, data collection procedure and training hyperparameters. Section \ref{sec:Results and discussions} shows the results and discussion about our algorithm performances and Section\ref{sec:conclusion} presents the conclusion and limitations.

%% file: Sections/drop_detection.tex
\begin{figure}[htbp]
      \centering
      \includegraphics[width=\columnwidth]{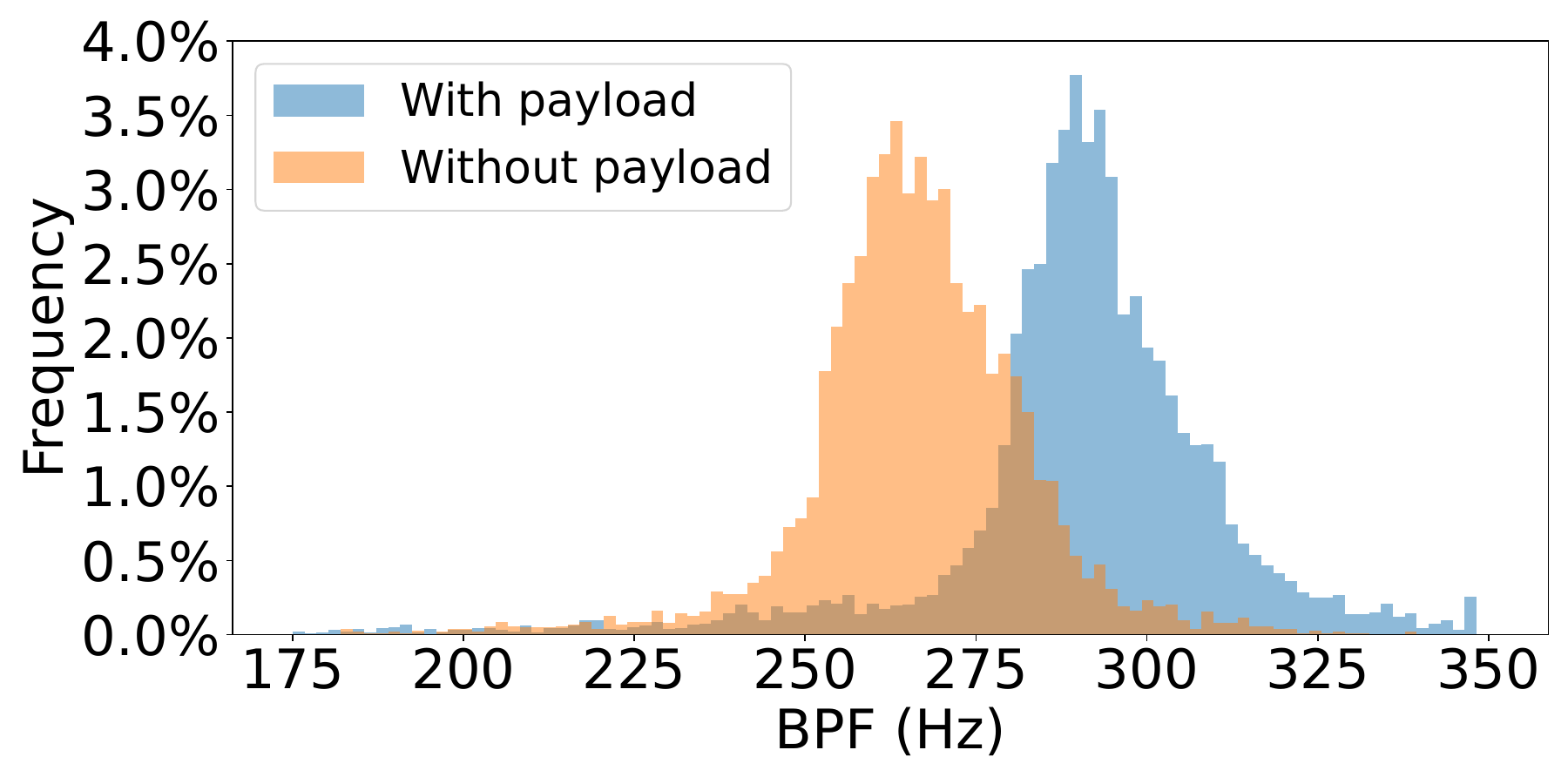}
      \caption{BPF distribution before and after a delivery. A payload shifts the distribution to the right as the effective drone weight increases. The ground truth BPF data from multiple delivery segments were used to generate this graph resulting in approximately 25 000 data points.}
      \label{fig:bpf distribution before after}
      \vspace{-10pt}
\end{figure}

Each propeller emits a sound characterized by the blade passing frequency (BPF) and some harmonics, defined as:
\begin{equation}
    BPF = N_p \omega,
\label{eq:fundamental propeller freq}
\end{equation}
where $N_p \in \mathbb{N}$ is the number of blades on the propeller and $\omega \in \mathbb{R}^+$ stands for the angular speed of the motor`s shaft in revolutions per second.
When a drone hovers, the speed of its motors depends on its weight and external perturbations such as wind. 
The BPFs of a drone are on average higher with a payload than without it. 
Figure \ref{fig:bpf distribution before after} shows this difference in the BPFs obtained from the control signal sent to each motor. 

\begin{figure}[htbp]
      \centering
      \includegraphics[width=\columnwidth]{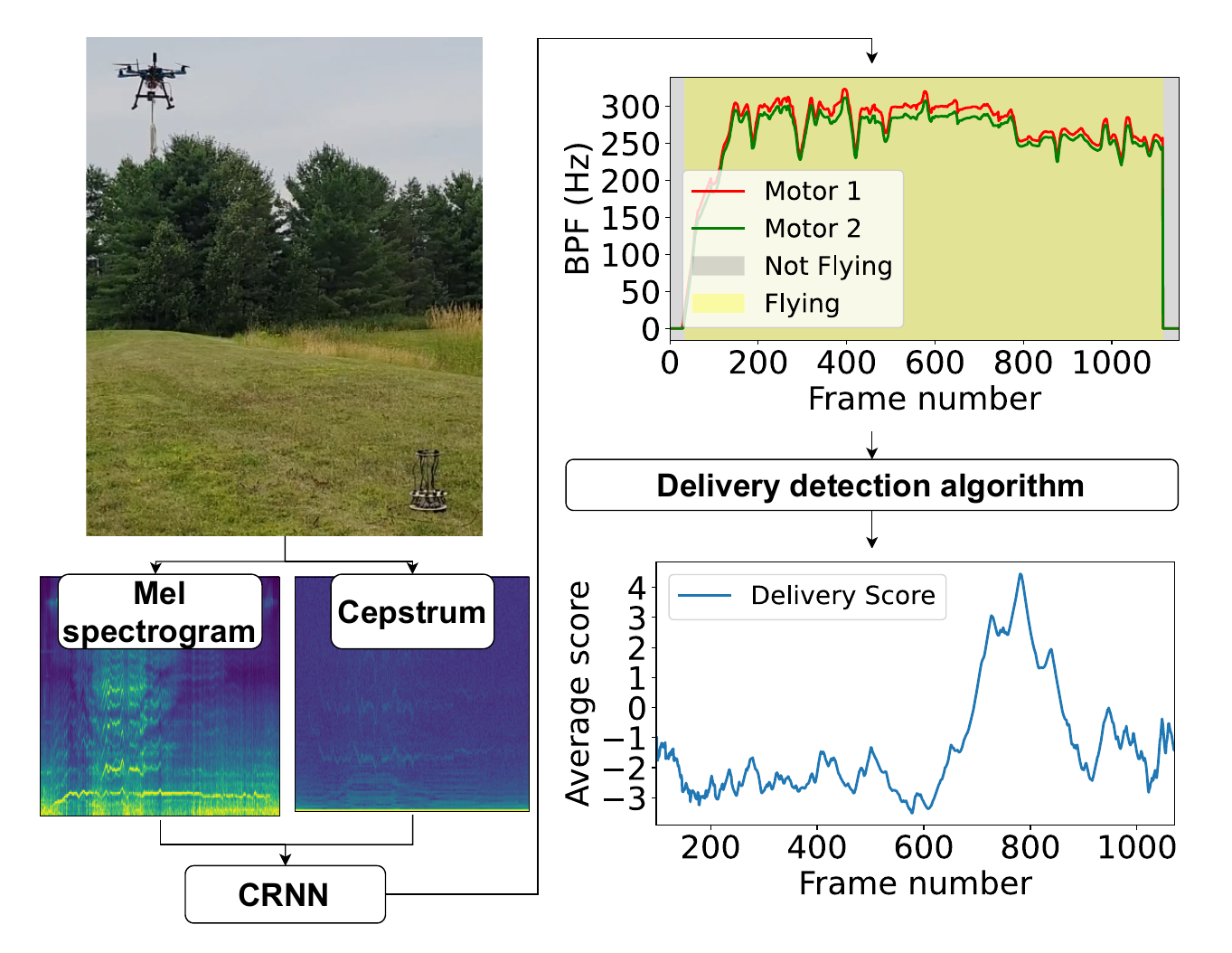}
      \caption{Architecture overview of the proposed algorithm.}
      \label{fig:architecture overview}
      \vspace{-10pt}
\end{figure}

To detect drone delivery event, we propose a two steps 
approach; 1) Use the acoustic signal recorded on the ground-based microphone array to detect the drone and extract the BPFs of two propellers using a convolutional and recurrent neural network (CRNN); 2) Use the estimated BPFs to identify possible delivery moments. The general approach is presented in Figure \ref{fig:architecture overview}.

\subsection{Blade passing frequency estimation and drone activity detection}
\label{sec:bpf estimation}

In practice, the BPF for each motor cannot be measured directly and must be estimated from the acoustic signals.
More specifically, the proposed algorithm estimates the BPFs of the two fastest motors, as these actuators are the most audible under various conditions.
A binary flag also predicts whether there is a drone sound event or not.
The input of the network is a 93 frames mel spectrogram generated using the parameters in Table \ref{tab:parameters mel} and a 93 frames power cepstrum which correspond to three seconds of audio. The higher quefrency bins of the power cepstrum were removed to match the dimension of the mel spectrogram.
Table \ref{tab:bpf estimator network} shows the architecture of the proposed supervised multitask CRNN, with two branches to predict the BPF and the drone activity.
The convolution blocks presented in Table \ref{tab:conv_block} extract the common features for both tasks, which are then fed to two branches, each with three stacked bi-directional gated recurrent unit (bGRU) and a two fully connected layer. The ReLU activation function ensures BPFs are positive, and the sigmoid function restricts the activity to the interval $[0,1]$. The BPFs are sorted from smallest to largest at each time frame to avoid ambiguity since they can overlap or cross during flights.

\begin{table}[htbp] 
\centering
\caption{Mel spectrogram parameters}
\begin{tabular}{ccccc}
\toprule
Window size & Hop size & Sampling & Nb of mel & Max freq \\
\midrule
2048        & 512      & 16 000 Hz   & 128       & 7000 Hz  \\
\bottomrule
\end{tabular}
\label{tab:parameters mel}
\end{table}
%% SI IL Y A DES MODIFICATION ICI, ME LE DIRE SVP PCQ LE TABLEAU VA PT SORTIR DES MARGES ET C'EST PAS OK POUR ICRA
\begin{table}
    \vspace{5pt}
    \centering
    \caption{Proposed CRNN architecture for BPF estimation and drone activity detection}
    \begin{tabular}{c|c|c|c}
    \toprule
        \multicolumn{4}{c}{Input (93 frames $\times$ 2 $\times$ 128 bins)}  \\
    \midrule
        ConvBlk(33) &  ConvBlk(21) & ConvBlk(11) & ConvBlk(3) \\
    \midrule
     
     \multicolumn{4}{c}{Concatenate ConvBlks =  (93 $\times$ 128 (channels) $\times$ 128 (bins))} \\
   
     \multicolumn{4}{c}{Flatten channels and bins = (93 $\times$ 16384)} \\
    
    \midrule
        \multicolumn{2}{c|}{bGRU($p=0.4$, $H=128$)} & \multicolumn{2}{c}{bGRU($p=0.4$, $H=128$)} \\
        \multicolumn{2}{c|}{bGRU($p=0.4$, $H=128$)} & \multicolumn{2}{c}{bGRU($p=0.4$, $H=128$)} \\
        \multicolumn{2}{c|}{bGRU($p=0.4$, $H=128$)} & \multicolumn{2}{c}{bGRU($p=0.4$, $H=128$)} \\
     
    \midrule
        
        \multicolumn{2}{c|}{Fully Connected($256$, $128$)} & \multicolumn{2}{c}{Fully Connected($256$, $128$)} \\
        \multicolumn{2}{c|}{ReLU} & \multicolumn{2}{c}{ReLU} \\
        
        \multicolumn{2}{c|}{Dropout($p=0.4$)} & \multicolumn{2}{c}{Dropout($p=0.4$)} \\
        \multicolumn{2}{c|}{Fully Connected($128$, $2$)} & \multicolumn{2}{c}{Fully Connected($128$, $1$)} \\
        \multicolumn{2}{c|}{ReLU} & \multicolumn{2}{c}{Sigmoid} \\
        
    \midrule
    
    \multicolumn{2}{c|}{BPF (93 frames $\times$ 2)} & \multicolumn{2}{c}{Activity (93 frames $\times$ 1)}\\
    \bottomrule
    \end{tabular}

    \vspace{5pt}
    $p$ = dropout probability, $H$ = hidden size
    \label{tab:bpf estimator network}
    \vspace{-10pt}
\end{table}

\begin{table}
    \centering
    \caption{Convolutional block definition}
    \begin{tabular}{c}
    
        ConvBlk (k) \\
    \midrule
        Input = (93 frames $\times$  2 $\times$ 128 bins) \\
    \midrule
         Conv2D($K = (3,k)$,$S = 1$,$P = (1,\floor{k/2})$,$C = 16$) +BN+ReLU\\
         Conv2D($K = (3,3)$,$S = 1$,$P = (1,1)$,$C = 32$)+BN+ReLU\\
    \midrule
    Output = (93 frames $\times$  32 channels $\times$ 128 bins) \\
    
    \bottomrule
    \end{tabular}
    
    \vspace{5pt}
    $K$ = kernel, $S$ = stride, $P$ = padding, $C$ = number of channels,
    \label{tab:conv_block}
    \vspace{-10pt}
\end{table}

%\begin{figure}[htbp]
%      \centering
%      \includegraphics[width=\columnwidth]{Photos/RPmAnnotationArch.jpg}
%      \caption{Proposed CRNN architecture for BPF estimation and drone detection}
%      \label{fig:bpf estimator network}
%\end{figure}

To train the multitask network, a combination of the mean squared (MSE) error loss function and the binary cross-entropy loss function (BCE) is used:
\begin{equation}
    L = \alpha (\hat{\vec{y}}_{b}-\vec{y}_{b})^2 + \beta (\vec{y}_{d} \log \hat{\vec{y}}_{d} + (1-\vec{y}_{d}) \log (1-\hat{\vec{y}}_{d})),
\label{eq:loss_function}
\end{equation}
where $\alpha, \beta \in \mathbb{R}^+$ are weights that balance both loss functions, $\vec{y}_{b}, \hat{\vec{y}}_{b} \in \mathbb{R}^+$ are the BPF label and prediction, respectively, and $\vec{y}_{d},\hat{\vec{y}}_{d} \in [0,1]$ are the drone activity label and prediction, respectively. In our case, $\alpha =1.0$ and $\beta=1.0$. 

\subsection{Delivery detection algorithm}
The delivery detection algorithm estimates, for each time frame $t$, the likelihood of a delivery and is based on a change point detection algorithm \cite{aminikhanghahi2017survey}. 
First we define the estimated BPF as $B^{m}_{t} \in \mathbb{R}^+$ where $m \in \{1,2\}$ is the motor index and $t \in [0,T-1]$ is the frame index. 
We also define a window that spans $n \in 2\mathbb{N}+1$ frames centered on the middle frame $t$ as:
\begin{equation}
    \vec{\gamma}^{m}_{n}|_t = \{ \mathbf{b}^m_n|_t,  \mathbf{a}^m_n|_t \},
\end{equation} \label{eq:window}
where
\begin{equation}\label{eq:split_window}
    \vec{b}^{m}_{n}|_t = \{ B^m_{t-\frac{n}{2}}, \dots, B^m_{t-1} \}, \vec{a}^{m}_{n}|_t = \{ B^m_{t}, \dots, B^m_{t+\frac{n}{2}} \}.
\end{equation}

Histograms $H_a$ and $H_b$ are computed for $\vec{a}^{m}_{n}|_t$ and $\vec{b}^{m}_{n}|_t$, respectively. 
The histograms are bounded between 100 Hz and 500 Hz and must contain a minimum of $n/4$ non-zero values to estimate the delivery score.
This prevents outliers and sequences with high rates of negative predictions from skewing the distribution. 
Samples for a time frame $t$ that fail to fulfill these criteria are discarded. 
The motor delivery score $s^m_n|_t$ is computed using a weighted sum (with weights $w_1$, $w_2$, $w_3$ and $w_4$) of the Chi-Squared distance $D_{\chi^2}$, the Jensen–Shannon divergence $D_{JSD}$, the histogram intersection score $D_{HI}$ and the difference in average between $H_a$ and $H_b$, denoted $\Delta\mu_{a,b}$:
\begin{equation}
     s^m_n|_t = w_1D_{\chi^2} + w_2D_{JSD} + w_3D_{HI} + w_4 \Delta\mu_{a,b}.
\end{equation}

\begin{figure}[ht]
      \centering
      \includegraphics[width=\columnwidth]{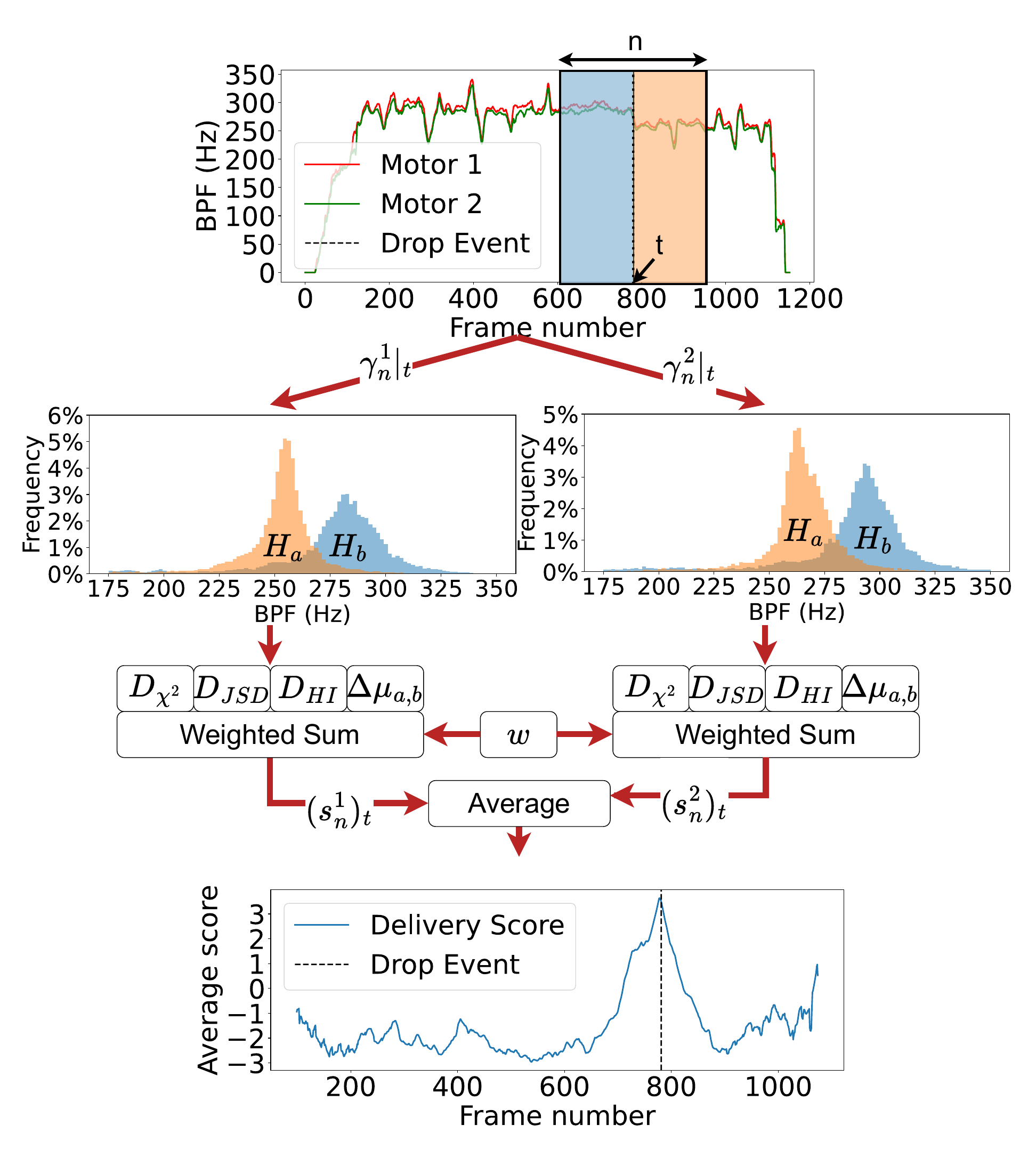}
      \caption{Delivery detection algorithm overview. $D_{\chi^2}$ is the Chi squared distance, $D_{JSD}$ the Jensen–Shannon divergence, $D_{HI}$ the intersection score and  $\Delta\mu_{a,b}$ the difference in average between both histograms}
      \label{fig:delivery detection algo}
      \vspace{-10pt}
\end{figure}

The final score $d_t$ is obtained by averaging the delivery score of both motors for all windows:
\begin{equation}
    d_t = \frac{1}{2N} \sum^N_{i=1} \left(s^1_{n_i}|_t + s^2_{n_i}|_t\right),
\end{equation}
where $N$ is the total number of windows and $n_i$ is the windows size for each window $i$. The weights $w_1=10,w_2=2,w_3=-20$ and $w_4=-0.05$ were chosen manually after testing different values. 
%The same process is repeated for the second motor BPFs and both delivery score are averaged resulting in the final delivery score for the frame $i$.
%within which histograms are computed for the first and second halves.  estimations to be considered "valid".  The discrepancy between the two histograms is then quantified using a weighted sum of the Chi-Squared distance, the Jensen divergence, the histogram intersection   %\cite{pearson_criterion_1900} \cite{lin_divergence_1991} \cite{swain_color_1991}and the difference in averages. A final score is obtained by averaging the discrepancy score of both motors and is assigned to the middle frame of the window .i.e where the split was made. 
%By computing the score over the $k$ time frame, d
Delivery events correspond to a time $t$ when the score $d_t$ exceeds a given threshold. 
In an online scenario, the windows size $n$ influences the responsiveness of the algorithm.
A shorter window provides a quicker response but can be affected by instant deviation while a larger window provides a slower but more precise response. 
We chose a total of 15 sliding windows ranging from 93 frames (3 seconds) to 465 frames (15 seconds). 
For every time frame, the score of each window is averaged with invalid estimation ignored. In a "real-time" scenario, the worst case lag introduced by our delivery detection algorithm is 7.5 seconds, which is half of the biggest window.
Figure \ref{fig:delivery detection algo} shows an overview of the delivery detection algorithm for one window size. 

%% file: Sections/experimental_setup.tex
% Experimental setup overview : Talk about drone (servo motor for drop, rpi for data capture, specs (weights, max payload). Talk about flights, what data is captured and how the data is organized, including the audio (MAYBE NOT INCLUDED). Talk about how much time flight were made
% Talk about MAVSDK and how the pwm is read.
%Talk about the microphone array (shape, photos, mic type, sr , etc)
% Talk about the flight environment (surounding noise, etc)
\subsection{Data collection}
\begin{figure}[htbp]
      \centering
      \includegraphics[width=\columnwidth]{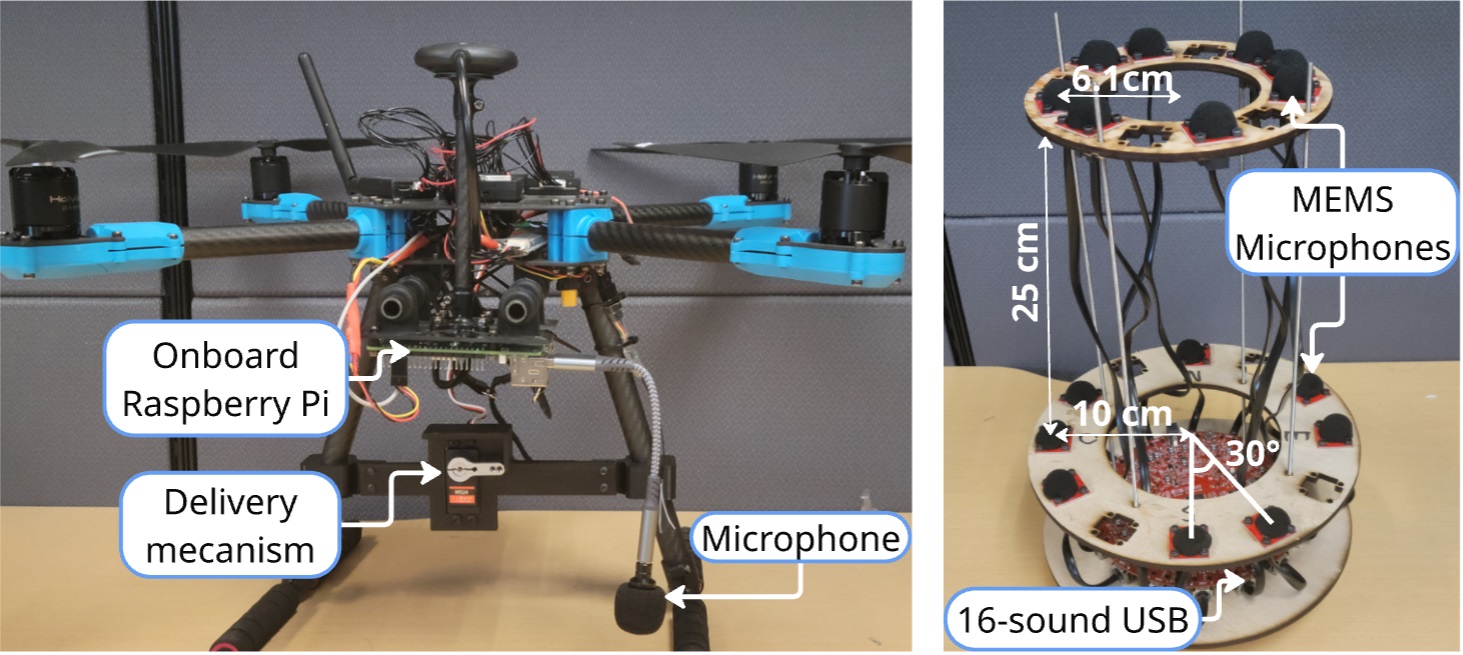}
      \caption{The modified Holybro X500 V2 drone and the 16 microphone array }
      \label{fig:drone + mic array annotated}
      \vspace{-5pt}
\end{figure}

A modified Holybro X500 V2 drone kit\footnote{\url{https://holybro.com/products/x500-v2-kits}} was used to conduct multiple flight tests.
The drone is equipped with a servo motor used for package delivery and an onboard Raspberry Pi (RPi) for telemetry and audio recording. 
A ground-based 16-microphone array (with micro-electromechanical systems (MEMS) microphones and the 16-sound USB card \cite{lagace2023ego}) captures audio data from the drone. 
The microphones are carefully positioned to minimize symmetry between the pairs, thereby reducing side lobes in localization algorithms. 
Figure \ref{fig:drone + mic array annotated} shows the drone and the microphone array setup used for data collection.

%The lower layer microphone holes are 10 cm from the center and spaced by 30 degrees. Top top level holes are 6.1 cm from it's center, spaced by 30 degrees but offset by 5 degrees from the bottom layer's hole. Both layers are spaced by 25 cm and microphones were placed to avoid symmetry.     

 During each flight, the RPi records acceleration, speed, orientation, GPS coordinates, servo position, motor Pulse-Width Modulation (PWM) signals, flight status and audio signals. 
 %The MAVSDK\footnote{\url{https://github.com/mavlink/MAVSDK}} library in C++ is used to retrieve telemetry data from the drone controller and the C Alsa Library\footnote{\url{https://www.alsa-project.org/alsa-doc/alsa-lib/}} was used to record the audio data. 
 %The RPi was connected via UART to the \textit{TELEM2} interface on the Pixhawk 6C flight controller. 
 To ensure data synchronization, each telemetry data point is saved along with its corresponding timestamp, and the start time of the audio recording is also logged. 
 Data synchronization is performed offline. 
%\footnote{\url{https://holybro.com/products/pixhawk-6c}}.
The audio from the 16-microphone array is recorded at a sample rate of 16,000 samples/sec on a dedicated laptop. 
Hand claps are captured at the beginning of each flight to manually synchronize the onboard and ground-based audio recordings. 

A total of 28 flights were recorded: 17 contains both onboard and array audio (143 minutes of flight time) and 11 contains only onboard audio (51 minutes of flight time).
Additionally, 39 minutes of background noise recordings were collected between flights, capturing sounds such as highway traffic, voices, hand claps and chainsaw.
Flight trajectories vary significantly, ranging from linear paths with small acceleration to complex maneuvers involving frequent changes in direction, altitude, and speed. 
The drone-to-array distance during these flights ranged from 0 to 160 m. 
Two additional flights were recorded with the drone hovering at various distances from the array, up to 180 meters. 
These flights were specifically designed to evaluate the performances of the CRNN-based BPF estimation across different distances.

A total of 23 delivery events were recorded, with a payload weighting between 500g and 600g. 
Six of these deliveries were recorded by both the onboard microphone and the microphone array, while the remaining 17 events were recorded only by the onboard microphone. 
For each delivery, 30 seconds of audio before and after the event were extracted and the BPFs were estimated using the network described in Section \ref{sec:bpf estimation and drop detection}. 
To prevent overfitting, the flights that contain delivery events were excluded from the training dataset. 
The recorded data are available here\footnote{\url{https://github.com/introlab/d3set}}.

%As outlined in \ref{sec:bpf estimation and drop detection}, one of the goal is to estimate the BPF of each motor thus the need for the ground truth BPF at each data point.

\begin{figure}[htbp]
      \centering
      \includegraphics[width=\columnwidth]{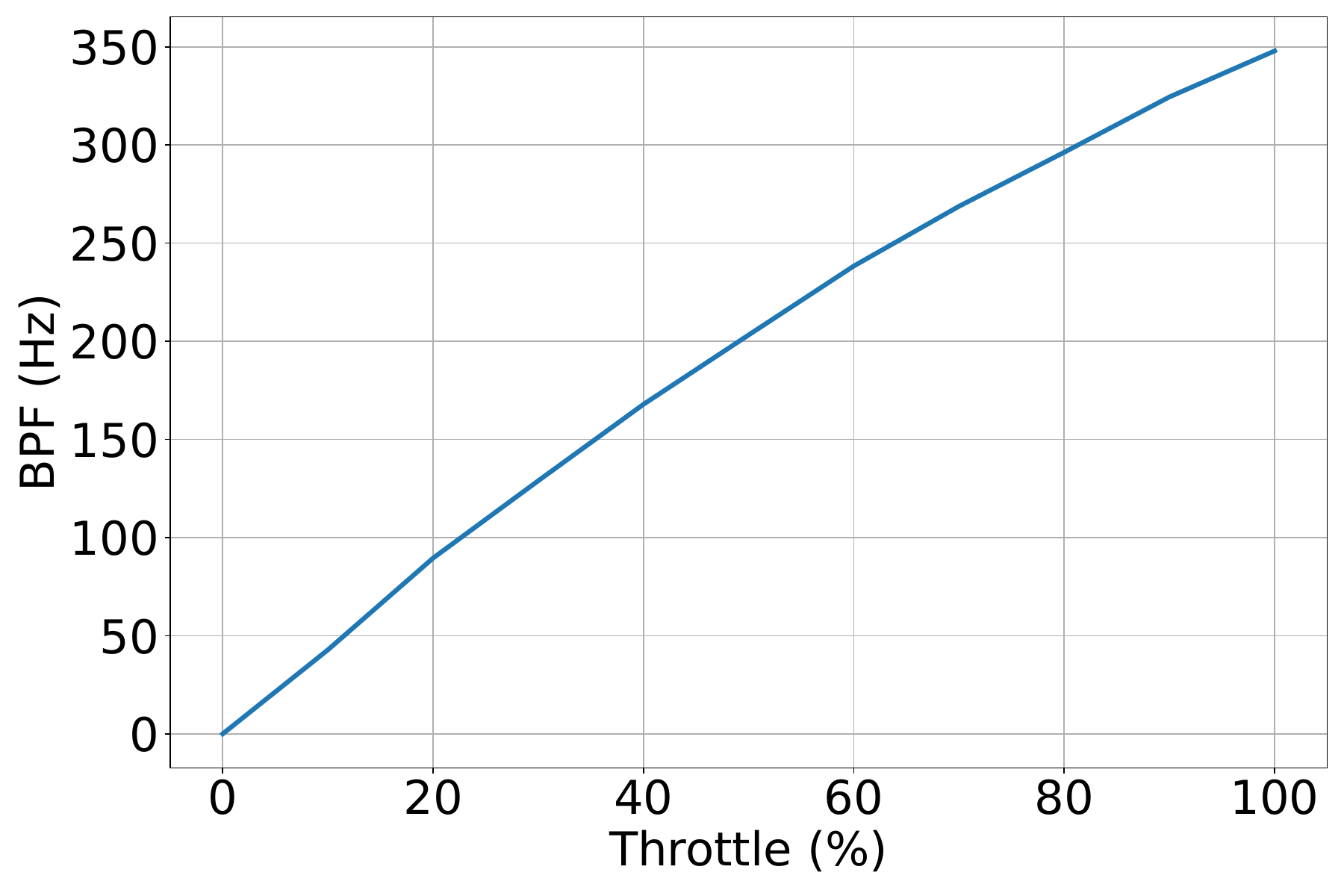}
      \caption{Measured BPF for PWM signal ranging from 1100 us (10\% throttle) to 2000 us (100\% throttle)}
      \label{fig:pwm to bpf curve}
      \vspace{-5pt}
\end{figure}

The motor's PWM signal, sent by the flight controller, is compared to the actual BPF of the propeller, which was measured using an optical tachometer and reflective tape placed on the propeller blades. 
BPF measurements were taken for each motor across a PWM range from 1100 $\mu$s (10\%) to 2000 $\mu$s (100\%) at 100 $\mu$s (10\%) intervals, with each measurement lasting between 3 to 7 seconds. 
These observations were averaged across all four motors and interpolated to generate the final PWM-to-BPF calibration curve shown in Figure \ref{fig:pwm to bpf curve}.

One important observation made during those measurements is that battery level affects the PWM-to-BPF relationship. The curve of the first motor (100\% battery) was slightly different from the last motor measurement(50\% battery). Because battery state was not recorded during flight, we had to a manual adjust the BPFs along every flight so they were aligned with the audible frequencies.
An offset in Hz was set for the beginning and the end of the recording and the BPF were gradually adjusted throughout the recording. In the majority of the recording, the maximal offset was around 20 Hz

\subsection{Data processing}
For the BPF estimation algorithm, only flights containing the onboard and ground-based array audio were used for the training, validation and testing. 
Out of the 17 available flights, 8 were used for the training, 3 for validation and 6 for testing. 
The test set was further split based on the recording device: one with the onboard audio and the other using the array audio. 
The BPFs were sorted from smallest to largest for each time frame to avoid ambiguity since BPFs from different motors can intersect. 
BPFs were also slightly smoothed using a Kalman filter to reduce abrupt variations that did not match the audible signal. Moreover, frame where the drone was more than 150 meters away from the microphone array were removed from the training and validation datasets.

Audio data augmentation was applied to the training dataset using the \textit{audiomentations}\footnote{\url{https://github.com/iver56/audiomentations}} library. 
The augmentation included: 1) light background noise with a SNR between 20 dB and 30 dB; 2) random gain from -5 dB to 5 dB; 3) time masking between 0.2 and 0.5 seconds; 4) frequency masking between 0 Hz and 7000 Hz.
%on the training set while the test set contains only clean audio samples. 
For each raw audio file, two augmented versions were generated by applying augmentations with randomized parameters every second. Each augmentation was applied with a probability of approximately 50\%.
AudioSet \cite{gemmeke2017audio} was used as background noise for augmentation and as complementary noise samples in the datasets. 
Additionally, we manually selected and incorporated noises such as helicopter, aircraft, mosquito, white noise and sinusoids ranging from 10 Hz to 6000 Hz in the training dataset. 
In the validation dataset, only noise from the AudioSet was added, while the test set included noise exclusively from the Urban16K \cite{salamon2014dataset} dataset.
Table \ref{table:dataset split} shows the composition of each dataset.

\begin{table}[htbp]
\caption{Datasets distribution in second}
\centering
\begin{tabular}{lrrrrr}
\toprule
Sound type & Train & Valid & Test & Onboard & Array \\
\midrule
Drone                           & 4752     & 2648  & 5993      & 2996              & 2996 \\
Drone augment                 & 9504     & 0     & 0      & 0                 & 0  \\
Added noise                     & 5709     & 1064 & 2997      & 0              & 0 \\
White noise                     & 570      & 0   & 0      & 0                 & 0  \\
Sinusoidal noise                & 570      & 0   & 0      & 0                 & 0 \\
Silence                         & 570      & 0   & 0      & 0                 & 0 \\
\midrule
Total                           & 21675    & 3712   & 8990    & 2996              & 2996  \\
\bottomrule
\end{tabular}
\label{table:dataset split}
\vspace{-15pt}
\end{table}

%% file: Sections/results_and_discussion.tex
\subsection{BPF estimator}
The peformance of the BPF estimator was evaluated using two metrics: 1) the mean absolute error (MAE) in Hz; 2) a custom metric called the masked MAE (MMAE).
The MMAE corresponds to the MAE  when the drone activity flag exceeds a predefined threshold. 
For drone activity detection, metrics include accuracy, recall and precision. Table \ref{table:metrics neural network} shows the metrics for the training, validation and both test sets.  

\begin{table}[h]
\centering
\caption{Neural network metrics for the training, validation and both test dataset }
\begin{tabular}{lrrrrr}
\toprule
Metrics         & Train  & Valid & Test   & Onboard   & Array     \\
\midrule
MAE (Hz)        & 3.9       & 7.6     & 13      & 6.2       & 31        \\
MMAE (Hz)       & 7.3       & 10.6    & 13      & 7.2       & 18         \\
Accuracy (\%)   & 99        & 97      & 97      & 99        & 91        \\
Precision (\%)  & 99        & 97      & 98      & 99        & 97        \\
Recall  (\%)    & 99        & 98      & 95      & 99        & 90        \\
\bottomrule
\end{tabular}
\label{table:metrics neural network}
\end{table}

Results with the onboard test set are promising with a drone activity accuracy, recall and precision of 99\%, and a BPF MAE of 6.2 Hz, which is close to the spectrogram frequency bin resolution of 7.8 Hz. 
The performances slightly degrades for the array test set, with a MAE of 31 Hz and an accuracy of 91\%. 
This is expected as the drone acoustic signal attenuates with increasing distance. 

\begin{figure*}[htbp]
    \vspace{10pt}
    \centering
    \begin{subfigure}[t]{.49\textwidth}
        \centering
        \includegraphics[width=\textwidth]{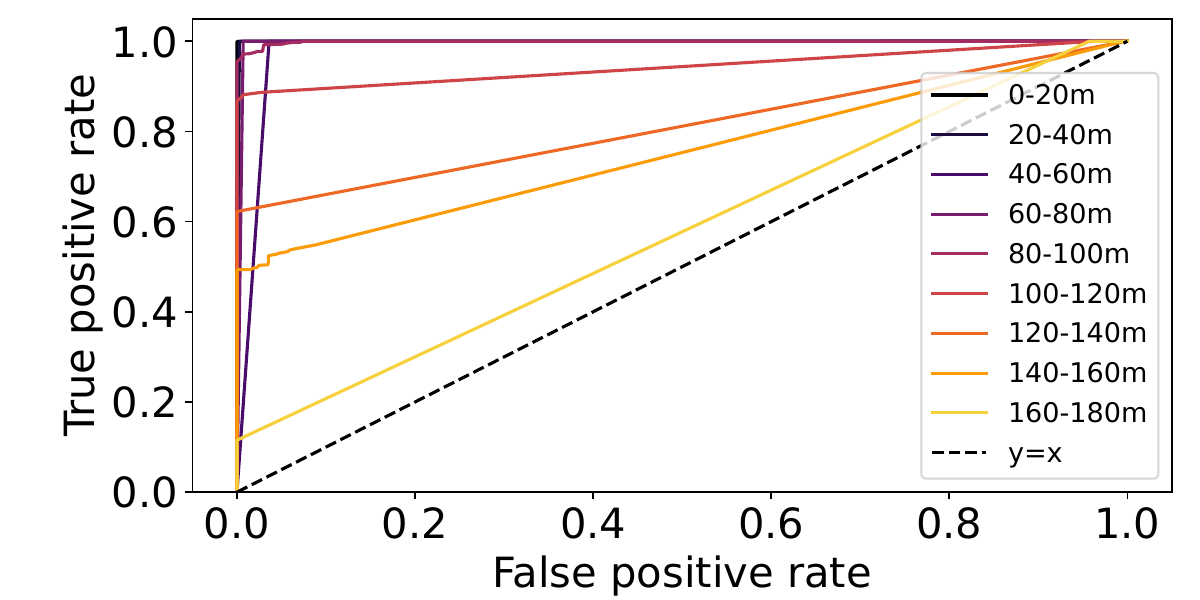}
        \caption{Drone activity detection ROC curves vs distances}
        \label{fig:roc curve metrics vs distance}
    \end{subfigure}%
    %%                          %   Don't leave blank line
    \hfill                      %%  <--- here:
    \begin{subfigure}[t]{.49\textwidth}
        \centering
        \includegraphics[width=\textwidth]{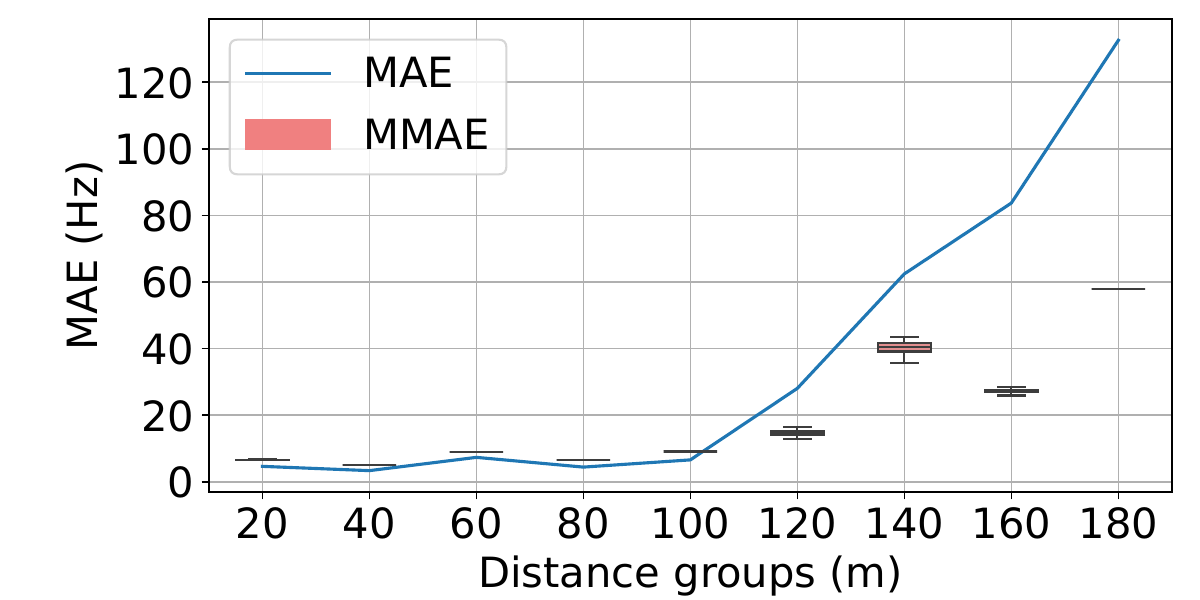}
        \caption{BPF MMAE and MAE vs distances}
        \label{fig:mae vs distance}
    \end{subfigure}
    \caption{Drone activity detection and BPF errors}
    \vspace{-5pt}
    \label{fig:performance vs distance}
\end{figure*}

\begin{figure*}[ht]
\centering
    \begin{subfigure}{.49\textwidth}
      \centering
      \includegraphics[width=0.9\textwidth]{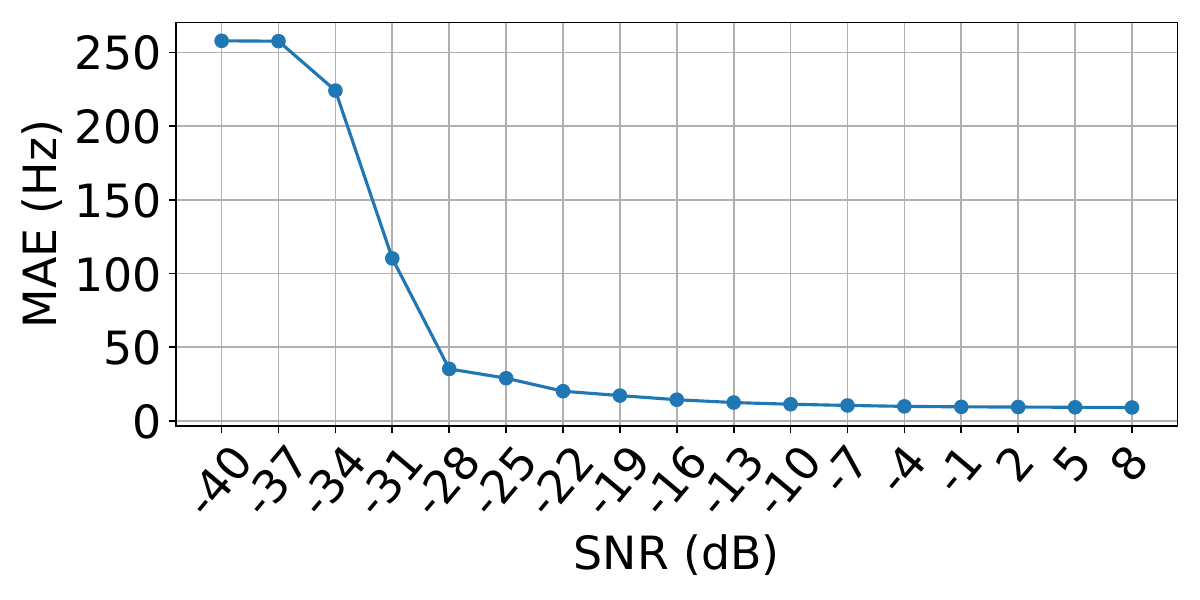}
      \caption{Urban noise}
      \label{fig:error vs snr background noise}
    \end{subfigure}%
    \begin{subfigure}{.49\textwidth}
      \centering
      \includegraphics[width=0.9\textwidth]{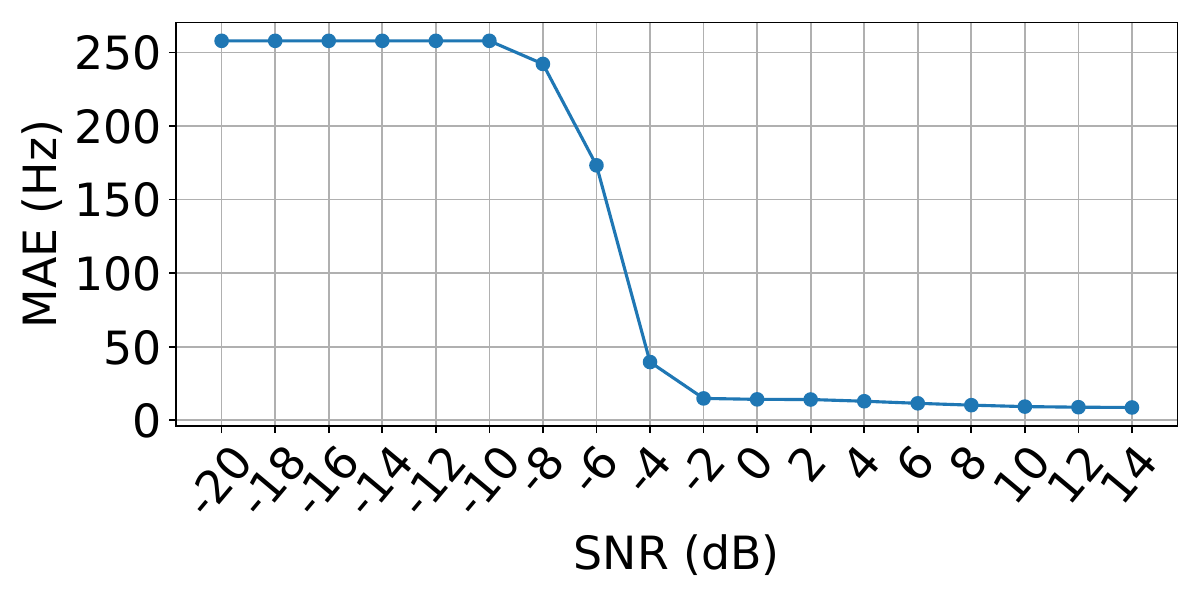}
      \caption{White noise}
      \label{fig:error vs snr white noise}
    \end{subfigure}
\caption{BPF MAE vs SNR}
\label{fig:both snr evaluation}
\vspace{-10pt}
\end{figure*}

The network performances were also evaluated across multiple drone-to-array distances and SNRs. 
For the distance-based evaluation, predictions from the excluded flight mentioned in Section \ref{sec:Experimental setup} were divided into 9 groups based on the drone-to-microphone array distance, ranging from 0 meter to 180 meters. 
In each group, random background noise was added to balance the groups with 50\% drone and 50\% noise labels. 
Figure \ref{fig:performance vs distance} shows the receiver operating characteristic (ROC) curves computed on the drone activity predictions, and the MAE and MMAE of the estimated BPF. 
Figure \ref{fig:both snr evaluation} shows the MAE with respect to the SNR. We used background urban noise recorded from the microphone array and white noise. 

%Insert table with metrics

%Since false negative largely skews the results, we only consider the predictions where the BPF are larger than 0 for the mean absolute distance metrics.
Figure \ref{fig:roc curve metrics vs distance} shows that the recall, or true positive rate for a fixed false positive rate decreases significantly with distance. 
In fact, the activity detection is accurate with a true positive rate of 100\% at a 10\% false positive rate up to 100 meters. 
Figure \ref{fig:mae vs distance} shows a similar pattern with the MAE staying below 15 Hz up to 100 meters. 
MMAE is generally lower than MAE across groups, which confirms that the activity flag can be used at inference to obtain more accurate predictions on the blade passing frequencies, especially at higher drone-to-array distances. 

Figure \ref{fig:error vs snr background noise} confirms that the BPF estimator is reliable from -22 dB onward. 
This robustness to background noise, which is predominantly concentrated in lower frequencies, is due to the presence of high-frequency components (around 5000 Hz) known as the azimuthal mode 2 vibration generated by the rotors \cite{henderson2018electric}. 
Figure \ref{fig:error vs snr white noise} confirms this observation as white noise also masks high frequencies, which results in an higher MAE at low SNRs. 

\subsection{Package delivery detection}
\begin{figure}[]
    \centering
\includegraphics[width=\columnwidth]{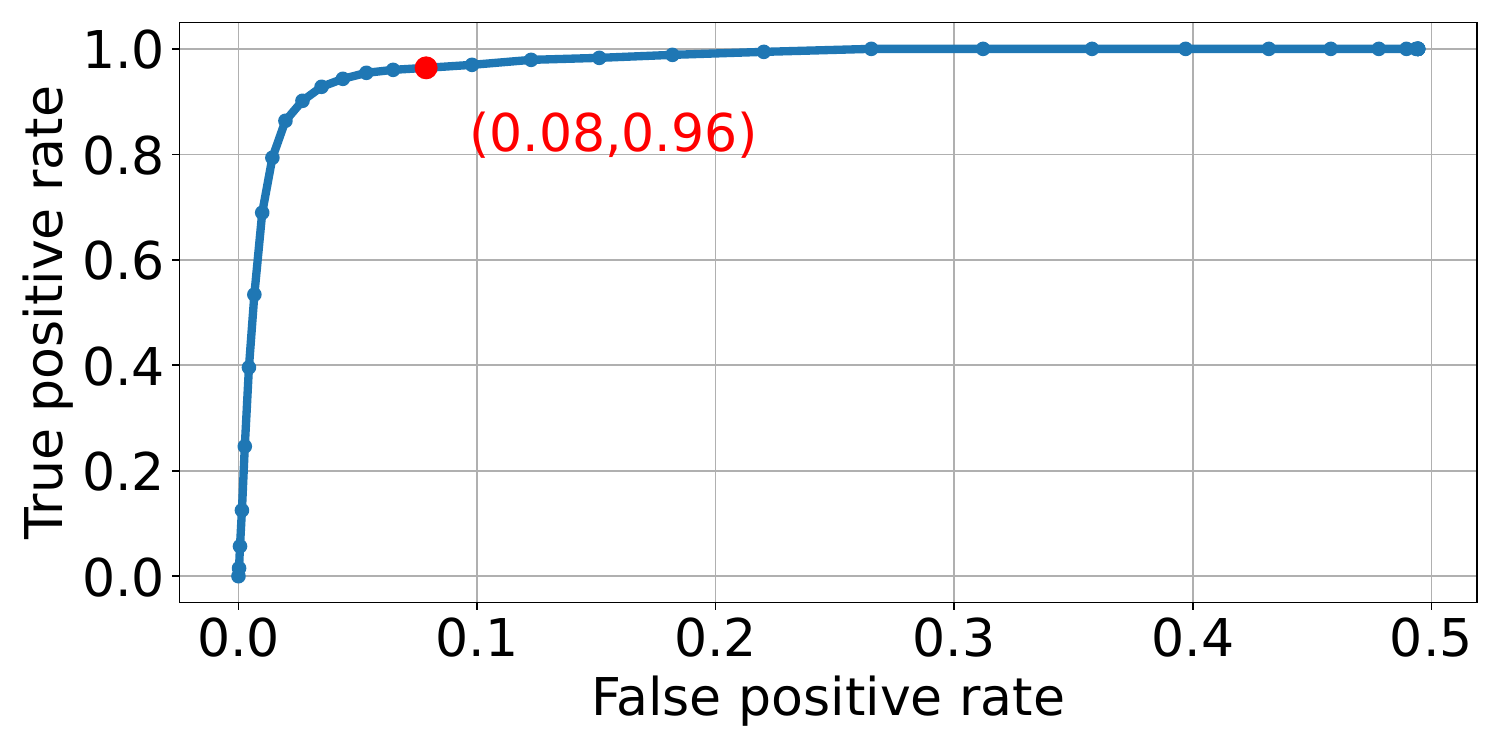}
    \caption{Delivery detection algorithm ROC curve}
    \label{fig:ROC drop detection}
    \vspace{-10pt}
\end{figure}

To evaluate the package delivery detection algorithm, each registered delivery event was extended to a 0.5 second window. 
This adjustment helped account for slight misalignments between the label and the event. 
For the 29 test cases, true positive rate and the false negative rate were computed across multiple threshold values, and the resulting ROC curve is presented in Figure \ref{fig:ROC drop detection}. 
At a false positive rate of 8\%, 96\% of delivery frames were identified correctly, demonstrating strong performance.
However, it is important to note that a relatively heavy payload (500g to 600g) was used, which caused significant changes in motor speeds at delivery. 
Smaller payloads would result in subtler speed variations, leading to smaller discrepancies between distributions and making detection more challenging. 
For now, the current event detection algorithm is agnostic of the drone position or speed, meaning that the Doppler effect, wind gust or intentional piloting maneuvers could lead to incorrect predictions.

%% file: Sections/conclusion.tex
This paper presents the first drone acoustic delivery detection system. 
Our method is based on a two steps approach; 1) Detect the drone and extract the BPFs of the two most audible propellers using a CRNN; 2) Identify possible delivery moments in the estimated BPFs sequence using a change point detection algorithm. Experimental results showed that the CRNN accurately estimated the BPFs with a mean absolute error of 16 Hz and estimated the drone presence with an accuracy of 97\% on realistic data with varying distances up to 100 m. We also conducted evaluation for multiples drone-to-array distances and SNRs. The distance evaluation highlighted that false negatives increases with distances. However, when a drone is detected, the estimated BPFs are accurate and we concluded that the proposed network is reliable up to 100 m. The SNR evaluation showed a great resilience to urban noise up to -22 dB SNR by utilizing drone's high frequency motor noise ($ \approx 5000 \text{Hz} $) to properly estimated the BPF even when the propeller fundamental frequencies are masked. The proposed delivery detection algorithm detected 96\% of the 29 delivery frames with a false positive rate of 8\%.

\textbf{Limitations}.
One major limitation is that only one drone was used to train the BPF estimation algorithm, which could make the method sensitive to domain mismatch and limits the general applicability.
Also, the delivery event detection approach relies solely on the differences between the data distribution besides an identified time frame. This difference is significant when using relatively heavy payloads compare to the drone weight ($ \approx 50\% $ ) but would be less significant when using lighter weight resulting in more false negatives for the same true positive rate. Moreover, since the delivery detection algorithm ignores the position of the drone, the Doppler effect, wind gusts or intentional drone maneuvers could lead to incorrect predictions. Direction of arrival estimation should be investigated in future work to make the system more robust.